# The Universal Approximation Power of Finite-Width Deep ReLU Networks

**Dmytro Perekrestenko**
Dept. IT & EE
ETH Zürich, Switzerland
pdmytro@nari.ee.ethz.ch

**Philipp Grohs**
Dept. Math.
University of Vienna, Austria
philipp.grohs@univie.ac.at

**Dennis Elbrächter**
Dept. Math.
University of Vienna, Austria
dennis.elbraechter@univie.ac.at

**Helmut Bölcskei**
Dept. IT & EE
ETH Zürich, Switzerland
boelcskei@nari.ee.ethz.ch

## Abstract

We show that finite-width deep ReLU neural networks yield rate-distortion optimal approximation [1] of polynomials, windowed sinusoidal functions, one-dimensional oscillatory textures, and the Weierstrass function, a fractal function which is continuous but nowhere differentiable. Together with their recently established universal approximation property of affine function systems [1], this shows that deep neural networks approximate vastly different signal structures generated by the affine group, the Weyl-Heisenberg group, or through warping, and even certain fractals, all with approximation error decaying exponentially in the number of neurons. We also prove that in the approximation of sufficiently smooth functions finite-width deep networks require strictly smaller connectivity than finite-depth wide networks.

## 1 Introduction

A theory establishing a link between the complexity of a neural network and the complexity of the function class to be approximated by the network was recently developed in [1]. Based on this theory, it was shown in [1] that all affine function classes are optimally representable by neural networks in the sense of Kolmogorov rate-distortion theory [4, 8]. Equivalently, this means that the approximation error decays exponentially in the number of neurons employed in the approximation.

The present paper explores the question of whether this universality result extends beyond affine function classes and answers it in the affirmative. Specifically, we consider the approximation of polynomials, windowed sinusoidal functions [7, 6], one-dimensional oscillatory textures according to [2], and the Weierstrass function, a fractal function which is continuous everywhere and differentiable nowhere. The central conclusion of this paper is that finite-width ReLU networks of depth scaling polylogarithmically in the inverse of the approximation error lead to exponentially decaying approximation error for all these different signal structures. This result is established by building on a recent breakthrough by Yarotsky [12] and recognizing that the width of networks approximating polynomials need not scale linearly in the degree of the polynomial as is the case in [12], but can actually be finite independently of the degree. This insight will also allow a sharp statement on the benefit of depth; specifically, we prove that in the approximation of sufficiently smooth functions finite-width deep networks require strictly smaller connectivity than finite-depth wide networks.

**Notation.** For the function $f(x)$ we define $\|f\|_{L^\infty[a,b]} := \inf\{C \geq 0 : |f(x)| \leq C,\text{ for all } x \in [a,b]\}$. We denote the set of all natural numbers in the interval $[1, N]$ by $[\![1, N]\!]$. Throughout the paper $\log$ stands for the logarithm to the base 2.

## 2 Setup and basic ReLU calculus

We start by defining ReLU neural networks.

**Definition 2.1.** *Let $L \geq 2, d, N_1, \ldots, N_L \in \mathbb{N}$. A map $\Phi : \mathbb{R}^d \to \mathbb{R}^{N_L}$ given by*

$$\Phi(x) = W_L(\rho\,(W_{L-1}(\rho\,(\ldots \rho\,(W_1(x)))))), \quad (1)$$

*with affine linear maps $W_\ell : \mathbb{R}^{N_{\ell-1}} \to \mathbb{R}^{N_\ell}, W_\ell(x) = A_\ell x + b_\ell, 1 \leq \ell \leq L, A_\ell \in \mathbb{R}^{N_\ell \times N_{\ell-1}}, b_\ell \in \mathbb{R}^{N_\ell}$, and the ReLU activation function $\rho(x) = \max(x, 0), x \in \mathbb{R}$, acting component-wise is called a ReLU neural network. Here $N_0 = d$ is the dimension of the input layer, $N_L$ denotes the dimension of the output layer, $L = \mathcal{L}(\Phi)$ stands for the number of layers (not counting the input layer) of the network, $N_1, \ldots, N_{L-1}$ are the dimensions of the $L - 1$ hidden layers, $M = \mathcal{M}(\Phi) := \max_\ell N_\ell$ denotes the network width, and the total number of nonzero weights in the network, i.e., the total number of nonzero entries in $A_\ell, b_\ell, \ell = 1, ..., N$, is referred to as network connectivity.*

We denote the class of ReLU networks $\Phi : \mathbb{R}^d \to \mathbb{R}^{N_L}$ with no more than $L$ layers, width no more than $M$, input dimension $d$, and output dimension $N_L$ by $\mathcal{NN}_{L,M,d,N_L}$. Note that the connectivity of $\Phi$ is upper-bounded by $LM(M + 1)$.

For later use we record two technical results. The first one shows how to augment network depth while retaining the network's input-output relation and without increasing its width.[1]

**Lemma 2.2.** *Let $K, L, M, d \in \mathbb{N}$ with $K > L$. Then, for every ReLU network $\Phi_1 \in \mathcal{NN}_{L,M,d,1}$, there exists a corresponding ReLU network $\Phi_2 \in \mathcal{NN}_{K,M,d,1}$ with $K$ layers such that $\Phi_2(x) = \Phi_1(x)$, for all $x \in \mathbb{R}^d$. Moreover, the weights of $\Phi_2$ consist of the weights of $\Phi_1$ and $\pm 1$'s.*

The next result formalizes the concept of a linear combination of networks.

**Lemma 2.3.** *Let $N, L_i, M_i, d_i \in \mathbb{N}, a_i \in \mathbb{R}, \Phi_i \in \mathcal{NN}_{L_i,M_i,d_i,1}, i = [\![1, N]\!], d = \sum_{i=1}^N d_i$. Then, there exist networks $\Phi^1 \in \mathcal{NN}_{L,M,d,N}$ and $\Phi^2 \in \mathcal{NN}_{L,M,d,1}$ with depth $L = \max_i L_i$ and width $M = \sum_{i=1}^N M_i$, where $\Phi^1$ realizes the map $x = (x_1^\top, x_2^\top, \ldots, x_N^\top)^\top \to (\Phi_1(x_1), \Phi_2(x_2), \ldots, \Phi_N(x_N))^\top, x \in \mathbb{R}^d$, and $\Phi^2$ realizes $x = (x_1^\top, x_2^\top, \ldots, x_N^\top)^\top \to \sum_{i=1}^N a_i \Phi_i(x_i), x \in \mathbb{R}^d$. Moreover, the weights of $\Phi^1$ consist of the weights of $\Phi_i, i = [\![1, N]\!]$, and $\pm 1$'s. The weights of $\Phi^2$ consist of the weights of $\Phi^1$ and $a_i, i = [\![1, N]\!]$.*

**Remark 2.4.** *Note that if the networks $\Phi_i$ in Lemma 2.3 have shared inputs, the resulting networks $\Phi^1$ and $\Phi^2$ will have fewer than $d = \sum_{i=1}^N d_i$ inputs.*

## 3 Approximation of polynomials

This section shows how the multiplication operation and polynomials can be approximated to within error $\epsilon$ with ReLU networks of finite width and of depth poly-logarithmic in $1/\epsilon$. We also note that the approximation results throughout the paper guarantee that the magnitude of the weights in the network does not grow faster than polynomially in the cardinality of the domain over which approximation takes place. Although not shown here for space constraints, the combination of finite width, depth scaling poly-logarithmically in $1/\epsilon$, and weights growing no faster than polynomially guarantees rate-distortion optimality in the sense of [1] and hence exponential error decay.

The proof ideas for the results in this section are inspired by [12] and the "sawtooth" construction of [11]. In contrast to [12], we consider networks without "skip connections" and of finite and explicitly specified width. Before starting with the approximation of $x^2$, we note that all our results apply to the multivariate case as well, but we restrict ourselves to the univariate case for simplicity of exposition.

---

[1] The proofs of the following two lemmata, along with all other proofs not provided in the paper, can be found in the supplement.

**Proposition 3.1.** *There exist constants $C, D > 0$ and networks $\Phi_\epsilon \in \mathcal{NN}_{L,5,1,1}$ with $L \leq C(\log(1/\epsilon))$, $\epsilon \to 0$, such that*

$$\|\Phi_\epsilon(x) - x^2\|_{L^\infty[0,1]} \leq \epsilon, \tag{2}$$

*and with the weights of $\Phi_\epsilon$ bounded (in absolute value) by $D$.*

*Proof.* Consider the "sawtooth" function $g : [0, 1] \to [0, 1]$,

$$g(x) = \begin{cases} 2x, & \text{if } x < \frac{1}{2}, \\ 2(1-x), & \text{if } x \geq \frac{1}{2}, \end{cases} \tag{3}$$

along with its $s$-fold composition

$$g_s = \underbrace{g \circ g \circ \cdots \circ g}_{s}, \quad s \geq 0, \tag{4}$$

where $g_0(x) := x$ and $g_1(x) := g(x)$. We next briefly review a fundamental result from [12] showing how the function $f(x) := x^2, x \in [0, 1]$, can be approximated by linear combinations of $s$-fold compositions $g_s$. Specifically, let $f_m$ be the piecewise linear interpolation of $f$ with $2^m + 1$ uniformly spaced "knots" according to

$$f_m\left(\frac{k}{2^m}\right) = \left(\frac{k}{2^m}\right)^2, \quad k = 0, \ldots, 2^m, \quad m \in \mathbb{N}_0.$$

The function $f_m$ approximates $f$ with error $\epsilon_m = 2^{-2m-2}$ in the sense of

$$\|f_m(x) - x^2\|_{L^\infty[0,1]} \leq 2^{-2m-2}.$$

Next, note that we can refine interpolations in the sense of going from $f_{m-1}$ to $f_m$ by adjustment with a sawtooth function according to

$$f_m(x) = f_{m-1}(x) - \frac{g_m(x)}{2^{2m}}. \tag{5}$$

This leads to the representation

$$f_m(x) = x - \sum_{s=1}^{m} \frac{g_s(x)}{2^{2s}}. \tag{6}$$

While Yarotsky's construction [12] is finalized by realizing (6) with the help of skip connections, we proceed by devising an equivalent fully connected network of width 5. As $g(x) = 2\rho(x) - 4\rho(x - 1/2) + 2\rho(x - 1)$, it follows that

$$g_m = 2\rho(g_{m-1}) - 4\rho(g_{m-1} - 1/2) + 2\rho(g_{m-1} - 1), \tag{7}$$

and consequently (5) can be rewritten as

$$f_m = \rho(f_{m-1}) - \rho(-f_{m-1}) - 2^{-2m}\Big(2\rho(g_{m-1}) - 4\rho(g_{m-1} - 1/2) + 2\rho(g_{m-1} - 1)\Big). \tag{8}$$

Equivalently, (7), (8) can be cast as a composition of affine linear maps and a ReLU nonlinearity according to

$$\begin{pmatrix} g_m \\ f_m \end{pmatrix} = W_1 \rho\left(W_2 \begin{pmatrix} g_{m-1} \\ f_{m-1} \end{pmatrix}\right), \tag{9}$$

with

$$W_1(x) = \begin{pmatrix} 2 & -2^{-2m+1} \\ -4 & 2^{-2m+2} \\ 2 & -2^{-2m+1} \\ 0 & 1 \\ 0 & -1 \end{pmatrix}^\top \begin{pmatrix} x_1 \\ x_2 \\ x_3 \\ x_4 \\ x_5 \end{pmatrix}, \quad W_2(x) = \begin{pmatrix} 1 & 0 \\ 1 & 0 \\ 1 & 0 \\ 0 & 1 \\ 0 & -1 \end{pmatrix} \begin{pmatrix} x_1 \\ x_2 \end{pmatrix} - \begin{pmatrix} 0 \\ 1/2 \\ 1 \\ 0 \\ 0 \end{pmatrix}. \tag{10}$$

Applying (9) iteratively initialized with $g_0(x) = x$, $f_0(x) = x$ shows that $f_m$ can be represented as a ReLU network in $\mathcal{NN}_{m+1,5,1,1}$ with weights uniformly bounded (in magnitude) by 4. With $\epsilon_m = 2^{-2m-2}$ and hence $\log(1/\epsilon_m) = 2m + 2$, the statement follows. □

With Proposition 3.1 we are now ready to show how ReLU networks can approximate the multiplication operation, which will then lead us to the approximation of arbitrary powers of $x$.

**Proposition 3.2.** *There exist a constant $C > 0$ and a polynomial $\pi$ such that for all $D \in \mathbb{R}_+$, there are networks $\Phi_{D,\epsilon} \in \mathcal{NN}_{L,15,2,1}$ with $L \leq C(\log(3\lceil D \rceil^2/\epsilon))$, $\epsilon \to 0$, such that*
$$\|\Phi_{D,\epsilon}(x,y) - xy\|_{L^\infty[-D,D]^2} \leq \epsilon, \tag{11}$$
*and with the weights of $\Phi_{D,\epsilon}$ bounded (in absolute value) by $|\pi(D)|$.*

The next result establishes that arbitrary polynomials can be approximated by ReLU networks of finite and explicitly specified width and depth growing logarithmically in the inverse of the approximation error. In particular, the width of the approximating network does not grow with the degree of the polynomial as is the case in [12], [3], [9]. This finite-width aspect is central to the approximation of sinusoidal functions by ReLU networks as described in the next section.

**Proposition 3.3.** *There exist a constant $C > 0$ and a polynomial $\pi$ such that for all $m \in \mathbb{N}$, $A \geq 1$, $p_m(x) = \sum_{i=0}^m a_i x^i$ with $\max_{i=1,\ldots,m} |a_i| \leq A$, and for all $D \in \mathbb{R}_+$, there are networks $\Phi_{p_m,D,\epsilon} \in \mathcal{NN}_{L,21,1,1}$ with $L \leq C\left(m\log(1/\epsilon) + m^2 \log(\lceil D \rceil) + m\log(m) + m\log(A) + m\right)$, $\epsilon \to 0$, such that*
$$\|\Phi_{p_m,D,\epsilon}(x) - p_m(x)\|_{L^\infty[-D,D]} \leq \epsilon, \tag{12}$$
*and where the weights of $\Phi_{p_m,D,\epsilon}$ are bounded (in absolute value) by $|\pi(D,A)|$.*

*Proof.* The proof will be effected by first showing that any monomial $x^k, k \geq 2$, can be approximated to within error $\epsilon$ by a composition of multiplication networks. Composing the networks corresponding to the individual monomials then completes the proof.

The network approximating $x^k, k \geq 2$, is constructed by iteratively composing multiplication networks as introduced in Proposition 3.2. The monomials $x^0$ and $x^1$ are realized through a two-layer network obtained by a simple affine transformation followed by the identity map $\rho(x) - \rho(-x)$. Next, let $0 \leq \eta \leq 1$, $H_1 := \lceil D \rceil$, $H_k := \lceil D \rceil^k + \eta \sum_{s=0}^{k-2} \lceil D \rceil^s, k \geq 2$, and define $\Psi_\eta^k$ recursively as $\Psi_\eta^0(x) := 1$, $\Psi_\eta^1(x) := x$, and $\Psi_\eta^k(x) = \Phi_{H_{k-1},\eta}(x, \Psi_\eta^{k-1}(x)), k \geq 2$, where $\Phi_{H_{k-1},\eta}$ is a multiplication network according to Proposition 3.2. We next show, by induction, that
$$\|\Psi_\eta^k(x) - x^k\|_{L^\infty[-D,D]} \leq \eta \sum_{s=0}^{k-2} \lceil D \rceil^s, \ k \geq 2. \tag{13}$$
The base case ($k=2$) follows from
$$\|\Psi_\eta^2(x) - x^2\|_{L^\infty[-D,D]} = \|\Phi_{H_1,\eta}(x,x) - x^2\|_{L^\infty[-D,D]} \leq \eta.$$
Next, we prove that (13) holds for $k \geq 3$, given that it holds for $k-1$. The induction assumption is
$$\|\Psi_\eta^{k-1}(x) - x^{k-1}\|_{L^\infty[-D,D]} \leq \eta \sum_{s=0}^{k-3} \lceil D \rceil^s, \text{ for } k \geq 3. \tag{14}$$
Now, thanks to $\|x\|_{L^\infty[-D,D]} \leq \lceil D \rceil \leq H_{k-1}$ and $\|\Psi_\eta^{k-1}\|_{L^\infty[-D,D]} \leq H_{k-1}$, which is by (14), we can apply Proposition 3.2 to conclude that $\Phi_{H_{k-1},\eta}(x, \Psi_\eta^{k-1}(x))$ approximates the product of $x$ and $\Psi_\eta^{k-1}(x)$ to within error $\eta$. It hence follows that
$$\|\Psi_\eta^k(x) - x^k\|_{L^\infty[-D,D]} \leq \|\Phi_{H_{k-1},\eta}(\Psi_\eta^{k-1}(x), x) - x\Psi_\eta^{k-1}(x)\|_{L^\infty[-D,D]}$$
$$+ \max_{[-D,D]} |x| \|\Psi_\eta^{k-1}(x) - x^{k-1}\|_{L^\infty[-D,D]} \leq \eta + \lceil D \rceil \eta \sum_{s=0}^{k-3} \lceil D \rceil^s = \eta \sum_{s=0}^{k-2} \lceil D \rceil^s,$$
which establishes the induction argument. In summary, we have shown that the network approximating $x^k$, for $k \geq 2$, is constructed according to $\Phi_{H_{k-1},\eta}(x, \Phi_{H_{k-2},\eta}(\ldots \Phi_{H_2,\eta}(x, \Phi_{H_1,\eta}(x,x))))$.

We are now ready to proceed to the construction of the network $\Phi_{D,\epsilon}$ approximating the polynomial $p_m(x) = \sum_{i=0}^m a_i x^i$. To this end, we note that the identity mapping $x \to x$ and the linear combination $(x,y)^\top \to x + a_{i-1}y$ can be realized as ReLU networks according to
$$\begin{pmatrix} 1 & -1 \end{pmatrix} \rho\left(\begin{pmatrix} 1 \\ -1 \end{pmatrix} x\right) \in \mathcal{NN}_{2,2,1,1},$$

and

$$(1 \quad -1 \quad a_{i-1} \quad -a_{i-1})\, \rho\left( \begin{pmatrix} 1 & 0 \\ -1 & 0 \\ 0 & 1 \\ 0 & -1 \end{pmatrix} \begin{pmatrix} x \\ y \end{pmatrix} \right) \in \mathcal{NN}_{2,4,2,1},$$

respectively. We shall also need the multiplication network $\Phi_{H_{i-1},\eta}(x,y) \in \mathcal{NN}_{L_i,15,2,1}$ with $L_i = \mathcal{O}(\log(1/\eta)) + C\log(H_{i-1}), \eta \to 0$, for some constant $C$. Then, thanks to Lemma 2.3, there exists a network $\Phi_i \in \mathcal{NN}_{L_i,21,3,3}$ with weights uniformly polynomially bounded in $D$ realizing the map

$$(x, s, y)^\top \to \left( x, s + a_{i-1}y, \Phi_{H_{i-1},\eta}(x,y) \right)^\top.$$

The network $\Phi_{D,\epsilon}$ approximating the polynomial $p_m(x) = \sum_{i=0}^m a_i x^i$ is now constructed as a composition of networks $\Phi_i, i = [\![2, m]\!]$, according to

$$\Phi_{D,\epsilon}(x) = (0 \quad 1 \quad a_m)\, \Phi_m\left( \Phi_{m-1}\left( \ldots \Phi_2\left( \begin{pmatrix} 1 \\ 0 \\ 1 \end{pmatrix} x + \begin{pmatrix} 0 \\ a_0 \\ 0 \end{pmatrix} \right) \right) \right) = \sum_{i=0}^m a_i \Psi_\eta^i(x).$$

The approximation error corresponding to the network $\Phi_{D,\epsilon}$ is obtained as

$$\left\| \Phi_{D,\epsilon}(x) - \sum_{i=0}^m a_i x^i \right\|_{L^\infty[-D,D]} \leq \sum_{i=2}^m |a_i| \left( \eta \sum_{s=0}^{i-2} \lceil D \rceil^s \right) \leq \eta \sum_{i=2}^m |a_i|(i-1)\lceil D \rceil^{i-2}$$

$$\leq \eta\, m \max_i |a_i| \sum_{i=2}^m \lceil D \rceil^{i-2} \leq A m^2 \lceil D \rceil^{m-2} \eta =: \epsilon,$$

where we applied the triangle inequality and used $\max_i |a_i| \leq A$, which is by assumption. Thanks to its compositional structure the width of $\Phi_{D,\epsilon}$ is equal to the width of the individual elements in the composition, namely 21, and its weights are uniformly polynomially bounded in $\lceil D \rceil$ as the elements in the composition all have weights that are uniformly polynomially bounded in $\lceil D \rceil$. The depth of $\Phi_{D,\epsilon}(x)$ is given by

$$L = \sum_{j=2}^m L_j = C_1(m-1)\log(1/\eta) + C_1 \sum_{i=1}^{m-1} \log(H_i) + C_2(m-1)$$

$$\leq C_1(m-1)\log(Am^2 \lceil D \rceil^m / \epsilon) + C_1 m \log(H_m) + C_2(m-1)$$

$$\leq C_1' m \log(1/\epsilon) + C_2' m^2 \log(\lceil D \rceil) + C_3' m \log(m) + C_4' m \log(A) + C_5' m,$$

where we used $H_m \leq m \lceil D \rceil^m$. This finalizes the proof. $\square$

We conclude this section with a result on ReLU networks approximating smooth functions with exponential accuracy. The proof of this statement, provided in the supplement, is based on the theory developed above.

**Definition 3.4.** For $D \in \mathbb{R}_+$, let the set $\mathcal{S}_D \subseteq C^\infty([-D,D],\mathbb{R})$ be given by

$$\mathcal{S}_D = \left\{ f \in C^\infty([-D,D],\mathbb{R}) \colon \|f^{(n)}(x)\|_{L^\infty[-D,D]} \leq n!, \text{ for all } n \in \mathbb{N}_0 \right\}. \tag{15}$$

**Lemma 3.5.** *There exist constants $C, C' > 0$ such that for all $D \in \mathbb{R}_+$, $f \in \mathcal{S}_D$, there exist networks $\Psi_{f,\epsilon} \in \mathcal{NN}_{L,28,1,1}$ with $L \leq C \lceil D \rceil (\log(1/\epsilon))^2, \epsilon \to 0$, satisfying*

$$\|\Psi_{f,\epsilon} - f\|_{L^\infty[-D,D]} \leq \epsilon, \tag{16}$$

*and where the weights of $\Psi_{f,\epsilon}$ are bounded (in absolute value) by $C' \max\{D, 1/D\}(1/\epsilon)$.*

Thanks to Proposition A.1, the weights of the networks $\Psi_{f,\epsilon}$ in Lemma 3.5 depending on $\epsilon$ does not constitute an issue with regards to exponential approximation accuracy, which requires the weights to be independent of $\epsilon$ and to scale at most polynomially in $D$. Specifically, Proposition A.1 allows to turn the networks $\Psi_{f,\epsilon}$ in Lemma 3.5 into finite-width networks with depth still scaling poly-logarithmically in $1/\epsilon$, weights bounded (in absolute value) by 2, and realizing the exact same function.

# 4 Approximation of sinusoidal functions

We are now ready to proceed to the approximation of sinusoidal functions.

**Theorem 4.1.** *There exist constants $C_1, C_2, C_3 > 0$ such that for all $D \geq 1, a \in \mathbb{R}_+$, there are networks $\Psi_{a,D,\epsilon} \in \mathcal{NN}_{L,21,1,1}$ with $L \leq C_1(\log(1/\epsilon))^2 + C_2 \log(\lceil aD \rceil), \epsilon \to 0$, such that*

$$\|\cos(a\cdot) - \Psi_{a,D,\epsilon}\|_{L^\infty([-D,D])} \leq \epsilon,$$

*and where the weights of $\Psi_{a,D,\epsilon}$ are bounded (in absolute value) by $C_3$.*

*Proof.* We start by noting that the MacLaurin series representation of $\cos$ is given by

$$\cos(x) = \sum_{n=0}^{\infty} \frac{(-1)^n}{(2n)!} x^{2n}, \quad \forall x \in \mathbb{R}.$$

Thanks to the Taylor theorem with remainder in Lagrange form, we have

$$\left\|\cos(2\pi x) - \sum_{n=0}^{N} \frac{(-1)^n}{(2n)!}(2\pi x)^{2n}\right\|_{L^\infty[0,1]} \leq \left\|\frac{(2\pi x)^{2N+1}}{(2N+1)!}\right\|_{L^\infty[0,1]} \leq \frac{(2\pi)^{2N+1}}{(2N+1)!}. \quad (17)$$

Next, we want to choose $N$ such that

$$\frac{(2\pi)^{2N+1}}{(2N+1)!} \leq \frac{\epsilon}{2}. \quad (18)$$

To this end, we start by noting that $n! \geq (\frac{n}{e})^n e$, which leads to

$$\frac{(2\pi)^{2N+1}}{(2N+1)!} \leq \frac{(2\pi)^{2N+1}}{(\frac{2N+1}{e})^{2N+1} e} \leq \left(\frac{2\pi e}{2N+1}\right)^{2N+1}.$$

Taking $N$ such that $2N + 1 = \lceil \ln(2/\epsilon) \rceil$ yields

$$\frac{(2\pi)^{2N+1}}{(2N+1)!} \leq \left(\frac{2\pi e}{\lceil \ln(2/\epsilon) \rceil}\right)^{\lceil \ln(2/\epsilon) \rceil} \leq \frac{\epsilon}{2}, \quad (19)$$

for $\epsilon \leq 2e^{-2\pi e^2}$. Next, applying Proposition 3.3 with $D = 1$ to $\sum_{n=0}^{N} \frac{(-1)^n}{(2n)!}(2\pi x)^{2n}$ and noting that $\max_n \left(\frac{(2\pi)^{2n}}{(2n)!}\right) \leq C' < \infty$, it follows that there is a network $\Phi_{\epsilon/2} \in \mathcal{NN}_{L,21,1,1}$ with $L \leq C_2(2N\log(1/\epsilon) + (2N)^2 \log(1) + 2N\log(2N) + 2N\log(C_1) + 1) \leq C_3(\log(1/\epsilon))^2, \epsilon \to 0$, such that

$$\left\|\sum_{n=0}^{N} \frac{(-1)^n}{(2n)!}(2\pi x)^{2n} - \Phi_{\epsilon/2}\right\|_{L^\infty[0,1]} \leq \left\|\sum_{n=0}^{N} \frac{(-1)^n}{(2n)!}(2\pi x)^{2n} - \Phi_{\epsilon/2}\right\|_{L^\infty[-1,1]} \leq \epsilon/2.$$

Moreover, the weights of this network are bounded by a uniform constant. Combining (17) and (19), it follows that the network $\Phi_{\epsilon/2}$ approximates $\cos(2\pi x)$ on $[0,1]$ to within accuracy $\epsilon$, i.e.,

$$\|\Phi_{\epsilon/2} - \cos(2\pi x)\|_{L^\infty[0,1]} \leq \epsilon. \quad (20)$$

We next extend this result to the approximation of $\cos(ax)$ on the interval $[0,1]$ for arbitrary $a > 0$. This will be accomplished by exploiting that $\cos(2\pi x)$ is 1-periodic and even. Specifically, for all $x \in [0,1]$, we have

$$\cos(2\pi g_s(x)) = \cos(2\pi 2^s x),$$

where $g_s$ is the "sawtooth" function from (4). Moreover, for every $a > 0$, there exists a $C_a \in (1/2, 1]$ such that $a/(2\pi) = C_a 2^{\lceil \log(a) - \log(2\pi) \rceil}$ and, therefore, for all $x \in [0,1]$, we have

$$\cos(ax) = \cos(2\pi 2^{\lceil \log(a) - \log(2\pi) \rceil} C_a x) = \cos\left(2\pi g_{\lceil \log(a) - \log(2\pi) \rceil}(C_a x)\right).$$

Upon noting that $g_{\lceil \log(a) - \log(2\pi) \rceil}(C_a x)$ takes value in $[0,1]$, it follows from (20) that

$$\left\|\Phi_{\epsilon/2}\left(g_{\lceil \log(a) - \log(2\pi) \rceil}(C_a x)\right) - \cos(ax)\right\|_{L^\infty[0,1]}$$
$$= \left\|\Phi_{\epsilon/2}\left(g_{\lceil \log(a) - \log(2\pi) \rceil}(C_a x)\right) - \cos\left(2\pi g_{\lceil \log(a) - \log(2\pi) \rceil}(C_a x)\right)\right\|_{L^\infty[0,1]} \leq \epsilon.$$

The network $\Psi_\epsilon(x) = \Phi_{\epsilon/2}\Big(g_{\lceil\log(a)-\log(2\pi)\rceil}(C_a x)\Big)$ therefore approximates $\cos(ax)$ on $[0,1]$ to within error $\epsilon$. Next, we note that $\Psi_\epsilon$ is the composition of two networks, namely $g_{\lceil\log(a)-\log(2\pi)\rceil}(C_a x)$, which is in $\mathcal{NN}_{L_1,3,1,1}$, $L_1 = \lceil\log(a)-\log(2\pi)\rceil + 1$, and $\Phi_{\epsilon/2} \in \mathcal{NN}_{L_2,21,1,1}$, $L_2 = \mathcal{O}((\log(1/\epsilon))^2)$, $\epsilon \to 0$. The width of $\Psi_\epsilon$ is therefore given by 21, its depth scales according to $L = \mathcal{O}((\log(1/\epsilon))^2) + \log(a)$, $\epsilon \to 0$, and its weights are uniformly bounded by a constant.

Finally, we extend the approximation of $\cos(ax)$ to intervals $[-D,D]$, for arbitrary $D \geq 1$. To this end, let $\tilde{a} > 0$, set $a = 4\pi\lceil\tilde{a}D\rceil$, and define the network $\tilde{\Psi}_{D,\epsilon}(x) := \Psi_\epsilon\Big(\frac{\tilde{a}x}{a} + \frac{1}{2}\Big)$, $x \in [-D,D]$. Since $\frac{\tilde{a}}{a} \leq 1/(4\pi D) \leq 1$, it follows that the network $\tilde{\Psi}_{D,\epsilon} \in \mathcal{NN}_{L,21,1,1}$ has weights uniformly bounded by a constant, and its depth satisfies $L = \mathcal{O}((\log(1/\epsilon))^2) + C_1 \log(\lceil\tilde{a}D\rceil)$, $\epsilon \to 0$. As $\Big(\frac{\tilde{a}x}{a} + \frac{1}{2}\Big) \in [0,1]$ for all $x \in [-D,D]$, the network $\tilde{\Psi}_{D,\epsilon}$ approximates $\cos(\tilde{a}x)$ on arbitrary intervals $[-D,D]$ thanks to

$$\Big\|\tilde{\Psi}_{D,\epsilon}(x) - \cos(\tilde{a}x)\Big\|_{L^\infty[-D,D]} = \Big\|\tilde{\Psi}_{D,\epsilon}(x) - \cos(\tilde{a}x + 2\pi\lceil\tilde{a}D\rceil)\Big\|_{L^\infty[-D,D]}$$

$$= \Big\|\Psi_\epsilon\Big(\frac{\tilde{a}x}{a} + \frac{1}{2}\Big) - \cos\Big(a\Big(\frac{\tilde{a}x}{a} + \frac{1}{2}\Big)\Big)\Big\|_{L^\infty[-D,D]} \leq \epsilon. \qquad \square$$

Local cosine bases [7] consist of time-frequency-shifted versions of windowed sinusoidal functions. It is therefore important to understand how time-shifted versions of $\cos(ax)$ can be realized through ReLU networks. Windowing will be dealt with in the next section.

**Corollary 4.2.** *There exist constants $C_1, C_2 > 0$ and a polynomial $\pi$ such that for all $a, b \in \mathbb{R}$, $a > 0, D \geq 1$, there are networks $\Phi^{\cos}_{D,\epsilon} \in \mathcal{NN}_{L,21,1,1}$ with $L \leq C_1((\log(1/\epsilon))^2) + C_2 \log(\lceil aD + |b|\rceil)$, $\epsilon \to 0$, such that*

$$\|\cos(a \cdot -b) - \Phi^{\cos}_{D,\epsilon}\|_{L^\infty([-D,D])} \leq \epsilon, \tag{21}$$

*and where the weights of $\Phi^{\cos}_{D,\epsilon}$ are bounded (in absolute value) by $|\pi(|b/a|)|$.*

**Remark 4.3.** *Approximation results for $\sin(ax-b)$ are obtained simply by noting that $\sin(x) = \cos(x - \pi/2)$ and applying Corollary 4.2.*

## 5 Oscillatory Textures and the Weierstrass Function

Consider the following function class consisting of one-dimensional "oscillatory textures" according to [2].

**Definition 5.1.** *Let the sets $\mathcal{F}_{D,a}$, $D, a \in \mathbb{R}_+$, be given by*

$$\mathcal{F}_{D,a} = \{\cos(ag)h : g, h \in \mathcal{S}_D\}. \tag{22}$$

The efficient approximation of functions in $\mathcal{F}_{D,a}$ with large parameters $a$ represents a notoriously difficult problem due to the combination of the rapidly oscillating cosine term and the warping $g$. The best available approximation results in the literature [2] are based on wave-atom dictionaries[2] and yield low-order polynomial approximation rates. In what follows we show that finite-width deep networks drastically improve these results to exponential approximation rates.

**Proposition 5.2.** *There exist constants $C, C', C'' > 0$ such that for all $D, a \in \mathbb{R}_+, f \in \mathcal{F}_{D,a}$, there are networks $\Phi_{f,\epsilon} \in \mathcal{NN}_{L,56,1,1}$ with $L \leq C\lceil D\rceil(\log(1/\epsilon))^2 + C'\log(\lceil aD\rceil)$, $\epsilon \to 0$, satisfying*

$$\|\Phi_{f,\epsilon} - f\|_{L^\infty[-D,D]} \leq \epsilon, \tag{23}$$

*and where the weights of $\Phi_{f,\epsilon}$ are bounded by $C''\max\{D, 1/D\}a/\epsilon$.*

Employing the same arguments as used after Lemma 3.5, the networks in Proposition 5.2 can be converted into networks with weights bounded by a uniform constant and hence realizing exponential approximation accuracy.

---

[2]To be precise, the results of [2] are concerned with the two-dimensional case, whereas we focus on the one-dimensional case. Note, however, that all our results can be readily extended to the multivariate case.

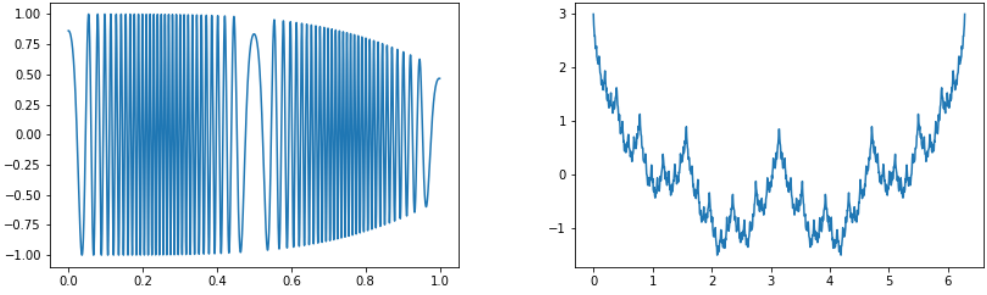

Figure 1: Left: A function in $\mathcal{F}_{1,100}$. Right: The function $W_{\frac{1}{\sqrt{2}},2}$.

**Remark 5.3.** *We note that Proposition 5.2 for $g(x) = x$ combined with Proposition 4.2 states that local cosine bases [7] can be approximated by deep ReLU networks with exponential error decay.*

We finally show how the Weierstrass function, a fractal function, which is continuous everywhere but differentiable nowhere can be approximated with exponential accuracy by deep ReLU networks. Specifically, we consider

$$W_{p,a}(x) = \sum_{k=0}^{\infty} p^k \cos(a^k \pi x), \quad \text{for } p \in (0, 1/2), \, a > 0. \tag{24}$$

Let $\alpha = -\frac{\log(p)}{\log(a)}$. It is well known [13] that $W_{p,a}$ possesses Hölder smoothness $\alpha$ which may be arbitrarily small, depending on $p$ and $a$, see Figure 1 right. While classical approximation methods are therefore not suitable, it turns out that deep finite-width networks achieve exponential approximation rate.

**Proposition 5.4.** *There exist constants $C_1, C_2, C_3, C_4 > 0$ such that, for all $p \in (0, 1/2)$, $a \in \mathbb{R}_+$, $D \geq 1$, there are networks $\Psi_{W_{p,a},D,\epsilon} \in \mathcal{NN}_{L,25,1,1}$ with $L \leq C_1(\log(1/\epsilon))^3 + C_2(\log(1/\epsilon))^2 \log(\lceil a \rceil) + C_3 \log(1/\epsilon) \log(D), \epsilon \to 0$, satisfying*

$$\|\Psi_{W_{p,a},D,\epsilon} - W_{p,a}\|_{L^\infty[-D,D]} \leq \epsilon, \tag{25}$$

*and where the weights of $\Psi_{W_{p,a},D,\epsilon}$ are bounded (in absolute value) by $C_4$.*

We conclude by mentioning a remarkable result bearing witness to the approximation power of deep ReLU networks in the context of PDEs. Specifically, it was shown recently in [10] that deep ReLU networks can approximate certain solution families of parametric PDEs depending on a large (possibly infinite) number of parameters while overcoming the curse of dimensionality.

## 6 Finite depth is not enough

We next show that finite-width deep networks require asymptotically—in $\epsilon \to 0$—smaller connectivity than finite-depth wide networks to approximate arbitrary periodic functions. This statement is then extended to sufficiently smooth non-periodic functions, thereby formalizing the benefit of deep networks over shallow networks for the approximation of a broad class of functions.

We start with preparatory material taken from [11].

**Definition 6.1** ([11]). *Let $k \in \mathbb{N}$. A function $f : \mathbb{R} \to \mathbb{R}$ is called $k$-sawtooth if it is piecewise linear with no more than $k$ pieces, i.e., its domain $\mathbb{R}$ can be partitioned into $k$ consecutive non-overlapping intervals so that $f$ is linear on each interval.*

**Lemma 6.2** ([11]). *Every $f \in \mathcal{NN}_{L,M,1,1}$ is $(2M)^L$-sawtooth.*

**Definition 6.3.** *For a non-constant $u$-periodic function $f \in C(\mathbb{R})$, we define*

$$\xi(f) := \inf_{c,d} \|f(x) - (cx + d)\|_{L^\infty[0,u]} > 0.$$

The quantity $\xi(f)$ measures the error incurred by the best linear approximation of $f$ on a segment of length equal to the period of $f$; $\xi(f)$ can hence be interpreted as quantifying the non-linearity of $f$. The next result states that finite-depth networks with width scaling poly-logarithmically in the highest frequency of the periodic function to be approximated can not achieve arbitrarily small approximation error.

**Proposition 6.4.** *Let $f \in C(\mathbb{R})$ be a non-constant $u$-periodic function. If $\xi(f) > 0$, then, for every polynomial $\pi$, there exists $a > 0$, such that for every network $\Phi \in \mathcal{NN}_{L,M,1,1}$ with L not depending on $a$ and $M = \pi(\log(a))$, it holds that*

$$\|f(a\cdot) - \Phi\|_{L^\infty[0,u]} \geq \xi(f).$$

Application of Proposition 6.4 shows that finite-depth networks, owing to $\xi(\cos) = 1$, require faster than poly-logarithmic growth of connectivity in $a$ to approximate $\cos(ax)$ with arbitrarily small error, whereas finite-width networks, thanks to Theorem 4.1, can accomplish this at logarithmic growth in connectivity. The next result, taken from [5], when combined with Lemma 6.2 allows us to extend this conclusion to more general functions.

**Theorem 6.5** ([5]). *Let $f \in C^3([a,b])$ and consider a piecewise linear approximation of $f$ on $[a,b]$ that is accurate to within $\epsilon$ in the $L^\infty[a,b]$-norm. The minimal number of linear pieces required to accomplish this scales according to*

$$s(\epsilon) \sim \frac{c}{\sqrt{\epsilon}}, \text{ where } c = \frac{1}{4}\int_a^b \sqrt{|f''(x)|}dx.$$

We are now ready to state our main result on depth-width tradeoffs for three-times continuously differentiable functions $f$.

**Theorem 6.6.** *Let $f \in C^3([a,b])$ and assume that $c := \frac{1}{4}\int_a^b \sqrt{|f''(x)|}dx > 0$. Then, for every polynomial $\pi$, there exists $\epsilon > 0$, such that for every network $\Phi \in \mathcal{NN}_{L,M,1,1}$ with L not depending on $\epsilon$ and $M = \pi(\log(1/\epsilon))$, it holds that*

$$\|f - \Phi\|_{L^\infty[a,b]} \geq \epsilon.$$

## A Trading off size of weights for depth

The following result shows how to trade off the size of weights in deep ReLU networks for depth.

**Proposition A.1.** *Let $L, M, d, N_L \in \mathbb{N}$, $\Phi \in \mathcal{NN}_{L,M,d,N_L}$ with weights bounded (in absolute value) by $B$. Then, there exists a network $\Psi$ satisfying*

1. $\Psi(x) = \Phi(x)$, for all $x \in \mathbb{R}^d$,
2. $\mathcal{M}(\Psi) = 2M(M+1)$,
3. $\mathcal{L}(\Psi) \leq (\lfloor \log(B) \rfloor + 3)L$, and
4. the weights of $\Psi$ are bounded (in absolute value) by $2$.

*Proof.* We start with the observation that for all $a \in \mathbb{R}$, there exists a network $\nu_a \in \mathcal{NN}_{\lfloor \log(|a|) \rfloor + 2, 2, 1, 1}$ with weights bounded (in absolute value) by $2$ and satisfying $\nu_a(x) = ax$, for all $x \in \mathbb{R}$. For $|a| > 2$ this network is given by

$$\nu_a(x) = \tfrac{a}{2^{\lfloor \log(|a|) \rfloor}} \begin{pmatrix} 1 & -1 \end{pmatrix} \rho\left(\begin{pmatrix} 2 & 0 \\ 0 & 2 \end{pmatrix} \rho\left(\ldots \rho\left(\begin{pmatrix} 2 & 0 \\ 0 & 2 \end{pmatrix} \rho\left(\begin{pmatrix} 1 \\ -1 \end{pmatrix} x\right)\right)\right)\right).$$

This implies that for every matrix $A = (a_{i,j}) \in \mathbb{R}^{N \times N'}$ with $\max_{i,j} |a_{i,j}| \leq |a|$, there exists a network $\alpha_A \in \mathcal{NN}_{\lfloor \log(|a|) \rfloor + 2, 2NN', N', NN'}$ with weights bounded (in absolute value) by $2$ and satisfying, for all $x \in \mathbb{R}^{N'}$,

$$\alpha_A(x) = (\nu_{a_{1,1}}(x_1), \nu_{a_{1,2}}(x_2), \ldots, \nu_{a_{1,N'}}(x_{N'}), \ldots, \nu_{a_{N,1}}(x_1), \nu_{a_{N,2}}(x_2), \ldots, \nu_{a_{N,N'}}(x_{N'}))^T$$
$$= (a_{1,1}x_1, a_{1,2}x_2, \ldots, a_{1,N'}x_{N'}, \ldots, a_{N,1}x_1, a_{N,2}x_2, \ldots, a_{N,N'}x_{N'})^T.$$

Similarly, observe that for all $b = (b_i) \in \mathbb{R}^N$ with $\max_i |b_i| \leq |a|$, there exists a network $\beta_b \in \mathcal{NN}_{\lfloor \log(|a|) \rfloor + 2, 2N, N', N}$ with weights bounded (in absolute value) by $2$ and satisfying $\beta_b(x) = b$, for all $x \in \mathbb{R}^{N'}$. Noting that $(Ax)_i = \sum_{j=1}^{N'} a_{i,j} x_j$, we can therefore conclude that for every $A = (a_{i,j}) \in \mathbb{R}^{N \times N'}$ and $b = (b_i) \in \mathbb{R}^N$ with $\max_{i,j} |a_{i,j}|, \max_i |b_i| \leq |a|$, there exist networks $\beta_{A,b} \in \mathcal{NN}_{\lfloor \log(|a|) \rfloor + 3, 2N(N'+1), N', N}$ and $\beta'_{A,b} \in \mathcal{NN}_{\lfloor \log(|a|) \rfloor + 2, 2N(N'+1), N', N}$ with weights bounded (in absolute value) by $2$ and satisfying, for all $x \in \mathbb{R}^{N'}$,

$$\beta_{A,b}(x) = \rho(Ax+b), \quad \beta'_{A,b}(x) = Ax+b.$$

We now apply this idea to construct a network equivalent to $\Phi$ in terms of the function realized but with weights bounded (in absolute value) by $2$. First, recall that

$$\Phi(x) = W_L(\rho(W_{L-1}(\ldots(W_1(x))))),$$

where $W_\ell(x) = A_\ell x + b_\ell$ with $A_\ell \in \mathbb{R}^{N_\ell \times N_{\ell-1}}$, $b_\ell \in \mathbb{R}^{N_\ell}$, $\ell = 1, 2, \ldots, L$. Let now, for $\ell \in \{1, 2, \ldots, L\}$, the network $\gamma_\ell$ be given by

$$\gamma_\ell = \begin{cases} \rho(W_\ell) & : \ell < L, \max_{i,j} |a_{i,j}^{(\ell)}|, \max_i |b_i^{(\ell)}| \leq 2 \\ \beta_{A_\ell, b_\ell} & : \ell < L, \max_{i,j} |a_{i,j}^{(\ell)}|, \max_i |b_i^{(\ell)}| > 2 \\ W_L & : \ell = L, \max_{i,j} |a_{i,j}^{(\ell)}|, \max_i |b_i^{(\ell)}| \leq 2 \\ \beta'_{A_L, b_L} & : \ell = L, \max_{i,j} |a_{i,j}^{(\ell)}|, \max_i |b_i^{(\ell)}| > 2, \end{cases} \quad (26)$$

and set

$$\Psi(x) = \gamma_L(\gamma_{L-1}(\ldots \gamma_2(\gamma_1(x)))).$$

By construction $\Psi(x) = \Phi(x)$, for all $x \in \mathbb{R}^d$, $\mathcal{M}(\Psi) = 2M(M+1)$, and the weights of $\Psi$ are bounded (in absolute value) by $2$. Further, note that the depth of each of the subnetworks $\beta_{A_\ell, b_\ell}, \ell \in \{1, 2, \ldots, L\}$, is bounded by $\lfloor \log(B) \rfloor + 3$, which implies $\mathcal{L}(\Psi) \leq (\lfloor \log(B) \rfloor + 3)L$. □

# B Proofs

## B.1 Proof of Lemma 2.2

*Proof.* We use $x = \rho(x) - \rho(-x)$, and note that, for $\Phi_1 \in \mathcal{NN}_{L,M,d,1}$,

$$\Phi_1'(x) = \rho(\Phi_1(x)) - \rho(-\Phi_1(x)) \in \mathcal{NN}_{L+1,M,d,1}, \quad (27)$$

with $\Phi_1'(x) = \Phi_1(x)$, for all $x \in \mathbb{R}^d$. The weights of $\Phi_1'(x)$ consist of the weights of $\Phi_1$ and $\{-1, +1\}$. $\Phi_2$ is obtained by applying the operation in (27) $K - L$ times. □

## B.2 Proof of Lemma 2.3

*Proof.* Let $k = \arg\max_i L_i$ and apply Lemma 2.2 with $K = L_k$ to the networks $\Phi_i, i = [\![1, N]\!]$, to get the networks $\tilde{\Phi}_i \in \mathcal{NN}_{L_k, M_i, d_i, 1}, i = [\![1, N]\!]$, satisfying $\tilde{\Phi}_i(x) = \Phi_i(x)$. The networks $\Phi^1(x) = \left(\tilde{\Phi}_1(x_1), \tilde{\Phi}_2(x_2), \ldots, \tilde{\Phi}_N(x_N)\right)^\top$ and $\Phi^2(x) = (a_1, a_2, \ldots, a_N)\Phi^1(x)$ of depth $L_k = \max_i L_i$ and width $M = \sum_{i=1}^N M_i$ satisfy the conditions of the statement. □

## B.3 Proof of Proposition 3.2

*Proof.* The proof is based on the identity

$$xy = \frac{1}{2}((x+y)^2 - x^2 - y^2), \quad (28)$$

which shows how multiplication can be implemented through the squaring operation. Let $\Psi_\delta(x) \in \mathcal{NN}_{L_0,5,1,1}$, $L_0 = \mathcal{O}(\log(1/\delta))$, $\delta \to 0$, be neural networks approximating $x^2$ according to Proposition 3.1, i.e., $\|\Psi_\delta(x) - x^2\|_{L^\infty[0,1]} \leq \delta$. We first extend Proposition 3.1 to an approximation result over the interval $[-D, D]$, with $D \in \mathbb{R}_+$. Specifically, we will need to realize $x^2, y^2, (x+y)^2$ over $[-D, D]$ and $[-D, D]^2$, respectively, through a neural network. This will be accomplished by first noting that

$$\left\|\lceil D \rceil^2 \Psi_\delta\left(\frac{|x|}{\lceil D \rceil}\right) - x^2\right\|_{L^\infty[-D,D]} \leq \lceil D \rceil^2 \delta, \quad (29)$$

and likewise

$$\left\|4\lceil D \rceil^2 \Psi_\delta\left(\frac{|x+y|}{2\lceil D \rceil}\right) - (x+y)^2\right\|_{L^\infty[-D,D]^2} \leq 4\lceil D \rceil^2 \delta. \quad (30)$$

The network $\Psi_\delta(|x|)$ has one layer more than $\Psi_\delta(x)$ as it implements $|x| = \rho(x) + \rho(-x)$ in the first layer. Next, defining the network

$$\Phi_{D,\epsilon}(x, y) = \frac{\lceil D \rceil^2}{2}\left(4\Psi_\delta\left(\frac{|x+y|}{2\lceil D \rceil}\right) - \Psi_\delta\left(\frac{|x|}{\lceil D \rceil}\right) - \Psi_\delta\left(\frac{|y|}{\lceil D \rceil}\right)\right),$$

it follows from Lemma 2.3 applied to $\Psi_\delta\left(\frac{|x+y|}{2\lceil D \rceil}\right) \in \mathcal{NN}_{L_0,5,2,1}$ and $\Psi_\delta\left(\frac{|x|}{\lceil D \rceil}\right), \Psi_\delta\left(\frac{|y|}{\lceil D \rceil}\right) \in \mathcal{NN}_{L_0,5,1,1}$ that $\Phi_{D,\epsilon}(x, y) \in \mathcal{NN}_{L,15,2,1}$, with $L = \mathcal{O}(\log(1/\delta))$. Using (28) in combination with (29) and (30), and applying the triangle inequality yields

$$\left\|\Phi_{D,\epsilon}(x,y) - \left(\frac{(x+y)^2}{2} - \frac{x^2}{2} - \frac{y^2}{2}\right)\right\|_{L^\infty[-D,D]^2} \leq 3\lceil D \rceil^2 \delta.$$

Taking $\delta = \frac{\epsilon}{3\lceil D \rceil^2}$, it follows that $\Phi_{D,\epsilon}(x, y)$ approximates the operation $xy$ on $[-D, D]^2$ to within error $\epsilon$. As $\Psi_\delta$ has weights uniformly bounded by a constant independent of $\delta$, the weights of $\Phi_{D,\epsilon}$ are uniformly bounded by a (quadratic) polynomial in $\lceil D \rceil$. The proof is concluded by noting that $L \leq C(\log(1/\delta)) = C(\log(3\lceil D \rceil^2/\epsilon)), \epsilon \to 0$.

□

### B.4 Proof of Lemma 3.5

*Proof.* We first consider the case $D = 1$. A fundamental result on Chebyshev interpolation, see e.g. [9, Lemma 3], guarantees, for all $f \in \mathcal{S}_1$, $n \in \mathbb{N}$, the existence of a polynomial $P_{f,n}$ of degree $n$ such that

$$\|f - P_{f,n}\|_{L^\infty[-1,1]} \leq \tfrac{1}{2^n(n+1)!}\|f^{(n+1)}\|_{L^\infty[-1,1]} \leq \tfrac{1}{2^n}. \tag{31}$$

Writing the polynomials $P_{f,n}$ as $P_{f,n} = \sum_{j=0}^n a_{f,n,j} x^j$, crude—but sufficient for our purposes— estimates show that there exists a constant $c > 0$ such that $\max_{j=1,\ldots,n}|a_{f,n,j}| \leq 2^{cn}$, for all $f \in \mathcal{S}_1$, $n \in \mathbb{N}$. Applying Proposition 3.3 to $P_{f,n}$ yields the existence of a constant $C > 0$ and networks $\Phi_{P_{f,n},1,\epsilon/2} \in \mathcal{NN}_{L,21,1,1}$ such that

$$\|\Phi_{P_{f,n},1,\epsilon/2} - P_{f,n}\|_{L^\infty[-1,1]} \leq \tfrac{\epsilon}{2}, \tag{32}$$

with

$$L \leq C(n\log(2/\epsilon) + n\log(n) + cn^2), \ \epsilon \to 0. \tag{33}$$

Setting $n = \lceil \log(2/\epsilon) \rceil$ and $\Psi_{f,\epsilon} = \Phi_{P_{f,n},1,\epsilon/2}$, and combining (31) and (32) establishes that for all $f \in \mathcal{S}_1$,

$$\|\Psi_{f,\epsilon} - f\|_{L^\infty[-1,1]} \leq \|\Psi_{f,\epsilon} - P_{f,n}\|_{L^\infty[-1,1]} + \|P_{f,n} - f\|_{L^\infty[-1,1]}$$
$$\leq \tfrac{\epsilon}{2} + \tfrac{1}{2^n} \leq \tfrac{\epsilon}{2} + \tfrac{\epsilon}{2} = \epsilon.$$

Note that for all $\epsilon \in (0, 1/2)$, we have $\lceil \log(2/\epsilon) \rceil \leq 2\log(2/\epsilon)$ and $\log(2/\epsilon) \leq 2\log(1/\epsilon)$. Hence (33) implies that for all $f \in \mathcal{S}_1$,

$$L \leq C(24 + 16\,c)(\log(1/\epsilon))^2 = C'(\log(1/\epsilon))^2, \ \epsilon \to 0. \tag{34}$$

Moreover, note that, thanks to Proposition 3.3, the $\Phi_{P_{f,n},1,\epsilon/2}$ have weights that are at most exponential in the degree $n$. As $n$ was taken to be logarithmic in $1/\epsilon$ this implies that the weights of $\Psi_{f,\epsilon}$ scale at most linearly in $1/\epsilon$. This completes the proof for the case $D = 1$.

We next prove the statement for $D \in (0, 1)$. To this end, we start by noting that for every $g \in \mathcal{S}_D$, with $D \in (0, 1)$, there exists an $f_g \in \mathcal{S}_1$ such that $f_g(x) = g(Dx)$, for all $x \in [-1, 1]$. Hence, there exists, for every $\epsilon \in (0, 1/2)$, a network $\Psi_{f_g,\epsilon} \in \mathcal{NN}_{L,21,1,1}$ with $L \leq C'(\log(1/\epsilon))^2$ satisfying $\sup_{x \in [-1,1]} |\Psi_{f_g,\epsilon}(x) - f_g(x)| \leq \epsilon$. The claim is established by noting that

$$\sup_{x \in [-D,D]} |\Psi_{f_g,\epsilon}(\tfrac{x}{D}) - g(x)| = \sup_{x \in [-D,D]} |\Psi_{f_g,\epsilon}(\tfrac{x}{D}) - f_g(\tfrac{x}{D})|$$
$$= \sup_{x \in [-1,1]} |\Psi_{f_g,\epsilon}(x) - f_g(x)| \leq \epsilon.$$

Owing to the scaling operation $\tfrac{x}{D}$, the weights of the approximating network $\Psi_{f_g,\epsilon}(\tfrac{x}{D})$ are upper-bounded by $C'(1/(\epsilon D))$, for some constant $C'$.

It remains to prove the statement for the case $D > 1$. This will be accomplished by approximating $f$ on intervals of length 2 (or less) and stitching the resulting approximations together using a localized partition of unity. To this end let $a, b \in \mathbb{R}$ such that $0 < b - a \leq 2$, and let $h \in C^\infty([a,b], \mathbb{R})$ with $\|h^{(n)}\|_{L^\infty[a,b]} \leq n!$, for all $n \in \mathbb{N}$. Next, note that for every $\epsilon \in (0, 1/2)$, there exists a network $\Psi'_{h,\epsilon} \in \mathcal{NN}_{L,21,1,1}$ with $L \leq C'(\log(1/\epsilon))^2$ such that $\sup_{x \in [-1,1]} |\Psi'_{h,\epsilon}(x) - h(x + \tfrac{a+b}{2})| \leq \epsilon$. Hence, the network $\Psi_{h,\epsilon} \in \mathcal{NN}_{L,21,1,1}$ given by $\Psi_{h,\epsilon}(x) = \Psi'_{h,\epsilon}(x - \tfrac{a+b}{2})$ satisfies

$$\sup_{x \in [a,b]} |\Psi_{h,\epsilon}(x) - h(x)| \leq \epsilon. \tag{35}$$

Next, define the intervals $I_{D,k} \subseteq \mathbb{R}$, $k \in \{-\lceil D \rceil + 1, \ldots, \lceil D \rceil - 1\}$, according to

$$I_{D,k} = \begin{cases} [k-1, k+1] & : -\lceil D \rceil + 1 < k < \lceil D \rceil - 1 \\ [-D, -\lceil D \rceil + 2] & : k = -\lceil D \rceil + 1 \\ [\lceil D \rceil - 2, D] & : k = \lceil D \rceil - 1. \end{cases}$$

By (35) it follows that, for all $D > 1$, $f \in \mathcal{S}_D$, $k \in \{-\lceil D \rceil + 1, \ldots, \lceil D \rceil - 1\}$, $\epsilon \in (0, 1/2)$, there exist networks $\Psi_{f,k,\epsilon}$ satisfying

$$\sup_{x \in I_{D,k}} |\Psi_{f,k,\epsilon}(x) - f(x)| \leq \tfrac{\epsilon}{4} \tag{36}$$

with $\mathcal{L}(\Psi_{f,k,\epsilon}) \leq C'(\log(4/\epsilon))^2$. We next build a partition of unity through neural networks. Specifically, let $\chi(x) = \rho(x+1) - 2\rho(x) + \rho(x-1)$ and $\chi_k \in \mathcal{NN}_{2,3,1,1}$, $k \in \mathbb{Z}$, with $\chi_k(x) = \chi(x-k)$. Note that

$$\sum_{k \in \mathbb{Z}} \chi_k(x) = 1, \quad \text{for all } x \in \mathbb{R}. \tag{37}$$

Define, for all $D > 1$, $f \in \mathcal{S}_D$, the neural network $\Psi_{f,\epsilon}$ as

$$\Psi_{f,\epsilon}(x) := \Phi_{2,\epsilon/4}(\chi_{-\lceil D \rceil}(x), \Psi_{f,-\lceil D \rceil+1,\epsilon}(x)) + \Phi_{2,\epsilon/4}(\chi_{\lceil D \rceil}(x), \Psi_{f,\lceil D \rceil-1,\epsilon}(x))$$
$$+ \sum_{k=-\lceil D \rceil+1}^{\lceil D \rceil-1} \Phi_{2,\epsilon/4}(\chi_k(x), \Psi_{f,k,\epsilon}(x)), \tag{38}$$

where $\Phi_{2,\epsilon/4}$ is the multiplication network from Proposition 3.2. Combining (36), (37), and Proposition 3.2 establishes that for all $D > 1$, $f \in \mathcal{S}_D$,

$$\|\Psi_{f,\epsilon} - f\|_{L^\infty[-D,D]} \leq \epsilon.$$

It remains to be shown that the networks $\Psi_{f,\epsilon}$, indeed, exhibit the desired properties. To this end, let $\alpha_{f,1,\epsilon}(x) := \Phi_{2,\epsilon/4}(\chi_{-\lceil D \rceil}(x), \Psi_{f,-\lceil D \rceil+1,\epsilon}(x))$, $\alpha_{f,2,\epsilon}(x) := \Phi_{2,\epsilon/4}(\chi_{\lceil D \rceil}(x), \Psi_{f,\lceil D \rceil-1,\epsilon}(x))$, and, for $k \in \{3, \ldots, 2\lceil D \rceil + 1\}$,

$$\alpha_{f,k,\epsilon}(x) := \Phi_{2,\epsilon/4}(\chi_{k-\lceil D \rceil-2}(x), \Psi_{f,k-\lceil D \rceil-2,\epsilon}(x)).$$

For $k \in \{1, \ldots, 2\lceil D \rceil + 1\}$, let

$$\beta_{f,k,\epsilon}(x_1, x_2, x_3) := \begin{pmatrix} x_1 \\ \alpha_{f,k,\epsilon}(x_2) \\ x_3 \end{pmatrix}, \quad \text{set} \quad \beta_0(x) := \begin{pmatrix} x \\ 0 \\ 0 \end{pmatrix},$$

and define $A \in \mathbb{R}^{3 \times 3}$ such that $A(y_1, y_2, y_3)^T = (y_1, y_1, y_2 + y_3)^T$, for all $y \in \mathbb{R}^3$. We can now rewrite (38) according to

$$\Psi_{f,\epsilon}(x) = (0 \ 1 \ 1) \beta_{f,2\lceil D \rceil+1,\epsilon}(A\beta_{f,2\lceil D \rceil,\epsilon}(\ldots(A\beta_{f,1,\epsilon}(A\beta_0(x))))).$$

Hence, $\Psi_{f,\epsilon}$ is a network of width 28 satisfying $\mathcal{L}(\Psi_{f,\epsilon}) \leq C'\lceil D \rceil (\log(1/\epsilon))^2$, $\epsilon \to 0$. Finally, note that, owing to the shift operation $x - (a+b)/2$ in (35), the weights of $\Psi_{f,\epsilon}$ are upper-bounded by $C'D(1/\epsilon)$, for some constant $C'$. This completes the proof. $\square$

### B.5 Proof of Corollary 4.2

*Proof.* Thanks to Theorem 4.1 there is a network $\Psi_{D,\epsilon}$ such that

$$\|\cos(a\cdot) - \Psi_{D,\epsilon}\|_{L^\infty([-D,D])} \leq \epsilon,$$

with weights uniformly bounded by a constant and $\mathcal{L}(\Psi_{D,\epsilon}) = \mathcal{O}(((\log(1/\epsilon))^2 + \log(\lceil aD \rceil))$, $\epsilon \to 0$. By Theorem 4.1 it hence follows that $\Phi_{D,\epsilon}^{\cos}(x) := \Psi_{E,\epsilon}(x - b/a)$ with $E := D + |b|/a$ satisfies (21). The network $\Phi_{D,\epsilon}^{\cos}$ is in $\mathcal{NN}_{L,21,1,1}$ with $L = \mathcal{O}((\log(1/\epsilon))^2) + \log(\lceil aD + |b| \rceil)$, $\epsilon \to 0$. The weights of $\Phi_{D,\epsilon}^{\cos}$ are bounded (in absolute value) by $|\pi(|b/a|)|$, for some polynomial $\pi$, owing to the shift operation $x - b/a$ in the first layer of the network. $\square$

### B.6 Proof of Proposition 5.2

*Proof.* For all $D, a \in \mathbb{R}_+$, $f \in \mathcal{F}_{D,a}$, let $g_f, h_f \in \mathcal{S}_D$ be given (implicitly) by $f = \cos(ag_f)h_f$. Note that Lemma 3.5 guarantees the existence of a constant $C > 0$ and of neural networks $\Psi_{h_f,\eta}, \Psi_{g_f,\eta} \in \mathcal{NN}_{L_1,28,1,1}$, $L_1 \leq C\lceil D \rceil (\log(1/\eta))^2$, $\eta \to 0$, such that for all $\eta \in (0, 1/2)$,

$$\|\Psi_{g_f,\eta} - g_f\|_{L^\infty[-D,D]} \leq \eta, \quad \|\Psi_{h_f,\eta} - h_f\|_{L^\infty[-D,D]} \leq \eta. \tag{39}$$

Theorem 4.1 further ensures the existence of constants $C_1, C_2 > 0$ and of neural networks $\Phi_{a,D,\epsilon} \in \mathcal{NN}_{L_2,21,1,1}$, $a > 0$, $\epsilon \in (0, 1/2)$, with $L_2 \leq C_1(\log(1/\epsilon))^2 + C_2 \log(\lceil aD \rceil)$, $\epsilon \to 0$, and

$$\|\cos(a \cdot) - \Phi_{a,D,\epsilon}\|_{L^\infty[-D,D]} \leq \tfrac{\epsilon}{3}. \tag{40}$$

Further, thanks to Proposition 3.2, there exists a constant $C'' > 0$ and neural networks $\mu_\epsilon \in \mathcal{NN}_{L_3,15,2,1}$, with $L_3 \leq C''(\log(1/\epsilon))$, $\epsilon \to 0$, such that

$$\sup_{x,y \in [-D,D]} |\mu_\epsilon(x,y) - xy| \leq \tfrac{\epsilon}{3}, \tag{41}$$

for all $\epsilon \in (0, 1/2)$. For all $D, a \in \mathbb{R}_+$, $f \in \mathcal{F}_{D,a}$, $\epsilon \in (0, 1/2)$, let $\eta := \frac{\epsilon}{12\lceil a \rceil}$ and define the neural networks

$$\Gamma_{f,\epsilon} = \mu_\epsilon(\Phi_{a,D,\epsilon}(\Psi_{g_f,\eta}), \Psi_{h_f,\eta}). \tag{42}$$

First, observe that (39) and (40) imply that

$$|\Phi_{a,D,\epsilon}(\Psi_{g_f,\eta}(x)) - \cos(a g_f(x))| \leq |\Phi_{a,D,\epsilon}(\Psi_{g_f,\eta}(x)) - \cos(a \Psi_{g_f,\eta}(x))|$$
$$+ |\cos(a \Psi_{g_f,\eta}(x)) - \cos(a g_f(x))| \leq \tfrac{\epsilon}{3} + a\eta,$$

for all $x \in [-D, D]$. Combining this with (39) and (41) yields

$$|\Gamma_{f,\epsilon}(x) - f(x)| = |\mu_\epsilon(\Phi_{a,D,\epsilon}(\Psi_{g_f,\eta}(x)), \Psi_{h_f,\eta}(x)) - \cos(a g_f(x)) h_f(x)|$$
$$\leq |\mu_\epsilon(\Phi_{a,D,\epsilon}(\Psi_{g_f,\eta}(x)), \Psi_{h_f,\eta}(x)) - \Phi_{a,D,\epsilon}(\Psi_{g_f,\eta}(x)) \Psi_{h_f,\eta}(x)|$$
$$+ |\Phi_{a,D,\epsilon}(\Psi_{g_f,\eta}(x)) \Psi_{h_f,\eta}(x) - \cos(a g_f(x)) h_f(x)|$$
$$\leq \tfrac{\epsilon}{3} + (\tfrac{\epsilon}{3} + a\eta) + \eta + (\tfrac{\epsilon}{3} + a\eta)\eta \leq \epsilon,$$

for all $x \in [-D, D]$, where we used $\|f\|_{L^\infty[-D,D]} \leq 1$, which is by definition. The width of $\Gamma_{f,\epsilon}$ is given by the sum of the widths of $\Psi_{h_f,\eta}$ and $\Psi_{g_f,\eta}$ as 56. Furthermore,

$$\mathcal{L}(\Gamma_{f,\epsilon}) = L_3 + L_2 + L_1 \leq C\lceil D \rceil (\log(1/\epsilon))^2 + C' \log(\lceil aD \rceil), \epsilon \to 0. \tag{43}$$

Finally, as the weights of the networks $\Phi_{a,D,\epsilon}, \mu_\epsilon$ are bounded by a constant (independent of $a$ and $\epsilon$) and the weights of $\Psi_{h_f,\eta}, \Psi_{g_f,\eta}$ are bounded by $C' \max\{D, 1/D\} \frac{1}{\eta}$, $\eta \to 0$, the weights of $\Gamma_{f,\epsilon}$ are bounded by $C'' \max\{D, 1/D\} \frac{a}{\epsilon}$, $\eta \to 0$. This completes the proof. $\square$

### B.7 Proof of Proposition 5.4

*Proof.* For $N \in \mathbb{N}$, $p \in (0, 1/2)$, $a \in \mathbb{R}_+$, $x \in \mathbb{R}$, let $S_{N,p,a}(x) = \sum_{k=0}^N p^k \cos(a^k \pi x)$. The geometric sum formula ensures that

$$|S_{N,p,a}(x) - W_{p,a}(x)| \leq \sum_{k=N+1}^\infty |p^k \cos(a^k \pi x)| \leq \sum_{k=N+1}^\infty p^k = \tfrac{1}{1-p} - \tfrac{1-p^{N+1}}{1-p} \leq 2^{-N}. \tag{44}$$

Let $N_\epsilon = \lceil \log(2/\epsilon) \rceil$, $\epsilon \in (0, 1/2)$. Next, note that Theorem 4.1 ensures the existence of constants $C_1, C_2 > 0$ and of networks $\phi_{D,\epsilon}^k \in \mathcal{NN}_{L_k,21,1,1}$, $a \in (0, \infty)$, $k \in \mathbb{N}_0$, $D \in [1, \infty)$, $\epsilon \in (0, 1/2)$, with $L_k \leq C_1(\log(N_\epsilon/\epsilon))^2 + C_2 \log(\lceil a^k \pi D \rceil)$, $\epsilon \to 0$, satisfying

$$\|\cos(a^k \pi \cdot) - \phi_{D,\epsilon}^k\|_{L^\infty[-D,D]} \leq \tfrac{\epsilon}{2(N_\epsilon+1)}. \tag{45}$$

Thanks to $x = \rho(x) - \rho(-x)$ and Lemma 2.2, there exists, for every $L \in \mathbb{N}$, a neural network $\tau_L \in \mathcal{NN}_{L,2,1,1}$ satisfying $\tau_L(x) = x$, for all $x \in \mathbb{R}$. For all $p \in (0, 1/2)$, $a \in (0, \infty)$, $k \in \mathbb{N}_0$, $D \in [1, \infty)$, $\epsilon \in (0, 1/2)$, define the neural networks

$$\psi_{D,\epsilon}^{p,a,0}(x) = \begin{pmatrix} x \\ p^0 \phi_{D,\epsilon}^0(x) \\ 0 \end{pmatrix} \quad \text{and} \quad \psi_{D,\epsilon}^{p,a,k}(x_1, x_2, x_3) = \begin{pmatrix} x_1 \\ p^k \phi_{D,\epsilon}^k(x_2) \\ x_3 \end{pmatrix}, k > 0, \tag{46}$$

and let $A \in \mathbb{R}^{3 \times 3}$ be such that $A(y_1, y_2, y_3)^T = (y_1, y_1, y_2 + y_3)^T$, for all $y \in \mathbb{R}^3$. Consider now, for $p \in (0, 1/2)$, $a \in (0, \infty)$, $D \in [1, \infty)$, $\epsilon \in (0, 1/2)$, the network $\Psi_{W_{p,a},D,\epsilon}$ defined according to

$$\Psi_{W_{p,a},D,\epsilon}(x) = \begin{pmatrix} 0 & 1 & 1 \end{pmatrix} \psi_{D,\epsilon}^{p,a,N_\epsilon}(A \psi_{D,\epsilon}^{p,a,N_\epsilon-1}(\ldots(A \psi_{D,\epsilon}^{p,a,0}(x)))). \tag{47}$$

Note that for all $p \in (0, 1/2)$, $a \in (0, \infty)$, $D \in [1, \infty)$, $\epsilon \in (0, 1/2)$, $x \in [-D, D]$,

$$|\Psi_{W_{p,a},D,\epsilon}(x) - S_{N_\epsilon,p,a}(x)| = \left| \sum_{k=0}^{N_\epsilon} p^k \phi_{D,\epsilon}^k(x) - \sum_{k=0}^{N_\epsilon} p^k \cos(a^k \pi x) \right|$$
$$\leq \sum_{k=0}^{N_\epsilon} p^k |\phi_{D,\epsilon}^k(x) - \cos(a^k \pi x)| \leq (N_\epsilon + 1) \tfrac{\epsilon}{2(N_\epsilon + 1)} = \tfrac{\epsilon}{2}.$$

Combining this with (44) establishes that for all $p \in (0, 1/2)$, $a \in (0, \infty)$, $D \in [1, \infty)$, $\epsilon \in (0, 1/2)$, $x \in [-D, D]$,

$$|\Psi_{W_{p,a},D,\epsilon}(x) - W_{p,a}(x)| \leq 2^{-\lceil \log(\frac{2}{\epsilon}) \rceil} + \tfrac{\epsilon}{2} \leq \tfrac{\epsilon}{2} + \tfrac{\epsilon}{2} = \epsilon.$$

The width of $\Psi_{W_{p,a},D,\epsilon}$ is 25, by construction, and its depth is given by

$$L = \sum_{k=0}^{N_\epsilon} L_k \leq (N_\epsilon + 1)C((\log(N_\epsilon/\epsilon))^2 + \log(\lceil a^{N_\epsilon} \pi D \rceil))$$
$$\leq C_1(\log(1/\epsilon))^3 + C_2(\log(1/\epsilon))^2 \log(\lceil a \rceil) + C_3 \log(1/\epsilon) \log(D).$$

Note that Theorem 4.1 guarantees that the weights of the $\phi_{D,\epsilon}^k$ are bounded by a constant (independent of $D$ and $\epsilon$) which implies that so are the weights of the $\Psi_{W_{p,a},D,\epsilon}$ This completes the proof.

□

### B.8 Proof of Proposition 6.4

*Proof.* For brevity we provide the proof for $a \in \mathbb{N}$ only and note that the general case follows by a simple modification. According to Lemma 6.2 all $\Phi \in \mathcal{NN}_{L,M,1,1}$ are $(2M)^L$-sawtooth. As $M = \pi(\log(a))$ by assumption, $\Phi \in \mathcal{NN}_{L,M,1,1}$ is $(2\pi(\log(a)))^L = \tilde{\pi}(\log(a))$-sawtooth, for some polynomial $\tilde{\pi}$. Since for every polynomial $\tilde{\pi}$, $\tilde{\pi}(\log(a)) = \mathcal{O}(a)$, $a \to \infty$, there exists $a_0 > 0$, such that $a > \tilde{\pi}(\log(a))$, for all $a \geq a_0$. Now, take any such $a \geq a_0$ and consider the $(u/a)$-periodic function $f(a\cdot)$ on $[0, u]$. The segment $[0, u]$ hence contains $a$ periods of $f(a\cdot)$. Since $\Phi$ is $\tilde{\pi}(\log(a))$-sawtooth, by assumption, it has no more than $\tilde{\pi}(\log(a))$ linear pieces on the segment $[0, u]$. Owing to $a > \tilde{\pi}(\log(a))$, it therefore follows that there exists a segment $[u_1, u_2] \subset [0, u]$ over which $\Phi$ is linear and that supports a full period of $f(a\cdot)$. We can therefore conclude that

$$\|f(a\cdot) - \Phi\|_{L^\infty[0,u]} \geq \|f(a\cdot) - \Phi\|_{L^\infty[u_1,u_2]} \geq \inf_{c,d} \|f(x) - (cx + d)\|_{L^\infty[0,u]} = \xi(f).$$

□

### B.9 Proof of Theorem 6.6

*Proof.* The proof will be effected by contradiction. Assume that the network $\Phi \in \mathcal{NN}_{L,M,1,1}$ approximates $f$ to within $\epsilon$ in the $L^\infty[a, b]$-norm. Thanks to Lemma 6.2, $\Phi \in \mathcal{NN}_{L,M,1,1}$ is $(2M)^L$-sawtooth and hence piecewise linear. This allows us to apply Theorem 6.5 to conclude that (the piecewise linear function) $\Phi$ approximating $f$ to within error $\epsilon$ has to be $\sim \epsilon^{-1/2}$-sawtooth. This, however, leads to a contradiction as $M = \pi(\log(1/\epsilon))$, which is by assumption, results, thanks to $L$ not depending on $\epsilon$, in $(2M)^L = \tilde{\pi}(\log(1/\epsilon))$ for some polynomial $\tilde{\pi}$ and $\tilde{\pi}(\log(1/\epsilon)) \in o(\epsilon^{-1/2})$, $\epsilon \to 0$.

□